\documentclass[letterpaper, 10 pt, conference]{ieeeconf}  
\IEEEoverridecommandlockouts                              

\usepackage{graphics} 
\usepackage{graphicx}
\usepackage{epsfig} 
\usepackage{mathptmx} 
\usepackage{times} 
\usepackage{amsmath} 
\usepackage{amssymb}  
\usepackage{cite}
\usepackage{blindtext}
\usepackage{color}
\usepackage[hidelinks]{hyperref}
\usepackage{array}
\usepackage{multirow}
\usepackage{diagbox}
\usepackage{makecell}
\usepackage{comment}
\title{\LARGE \bf
Spot-Compose: A Framework for Open-Vocabulary Object Retrieval and Drawer Manipulation in Point Clouds
}

\author{Oliver Lemke$^{*}$, Zuria Bauer, René Zurbrügg, Marc Pollefeys, Francis Engelmann$^{\dagger}$, and Hermann Blum$^{\dagger}$ %
\thanks{$^*$Corresponding Author: Oliver Lemke $<$olemke@ethz.ch$>$}%
\thanks{$^{\dagger}$Equal supervision.
This research is partially supported by the ETH AI Center and by a Career
Seed Awards funded by the ETH Zurich Foundation.}
}

\begin{document}
\newcommand{\blue}[1]{\color{blue}#1\color{black}}
\maketitle

\thispagestyle{empty}
\pagestyle{empty}


\begin{abstract}
In recent years, modern techniques in deep learning and large-scale datasets have led to impressive progress in 3D instance segmentation, grasp pose estimation, and robotics.
This allows for accurate detection directly in 3D scenes, object- and environment-aware grasp prediction, as well as robust and repeatable robotic manipulation.
This work aims to integrate these recent methods into a comprehensive framework for robotic interaction and manipulation in human-centric environments.
Specifically, we leverage 3D reconstructions from a commodity 3D scanner for open-vocabulary instance segmentation, alongside grasp pose estimation, to demonstrate dynamic picking of objects, and opening of drawers.
We show the performance and robustness of our model in two sets of real-world experiments including dynamic object retrieval and drawer opening, reporting a 51\% and 82\% success rate respectively.
Code of our framework as well as videos are available on: \blue{\href{https://spot-compose.github.io/}{https://spot-compose.github.io/}}.

\end{abstract}

\section{INTRODUCTION}

One of the pinnacle achievements in the field of robotics is to develop systems capable of understanding and navigating spaces designed for humans.
This task poses significant challenges due to the high variability and complexity of human-centric environments, requiring good semantic understanding and precise manipulation capabilities.
Recent advancements in 3D scanning technologies, perception models, and intricate manipulation algorithms have collectively facilitated a leap in robotic abilities, enabling more nuanced and effective interactions within everyday human spaces.
This work introduces a framework that utilizes these advancements, while leveraging modern models for instance segmentation and grasp pose estimation using the Boston Dynamics Spot robot.
The key contributions of this paper include:

\begin{itemize}
    \item A modular framework on top of the Spot SDK, providing a flexible platform for the integration of modern machine perception techniques. 
    \item Integration of advanced models for perception and grasp estimation, enabling versatile interactions with diverse objects in human-centric environments.
    \item Real-world experiments, including dynamic object retrieval and drawer manipulation tasks.
\end{itemize}

\section{RELATED WORK}
\begin{figure*}
    \centering
    \includegraphics[width=\linewidth]{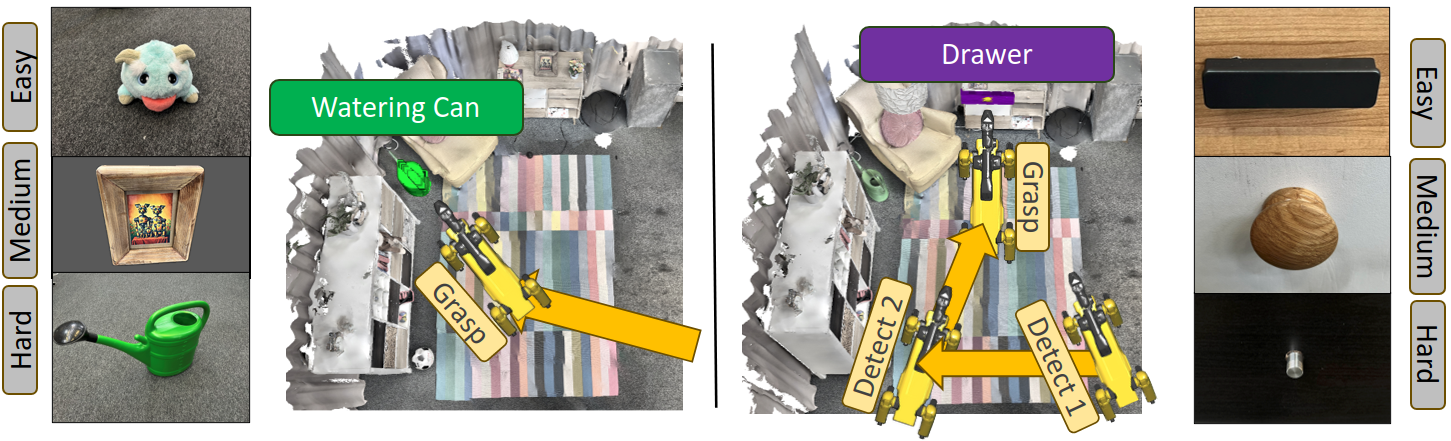}
    \caption{\textbf{Adaptive grasping and drawer interaction   pipeline.}
    On the left, we illustrate the grasping sequence initiated by the successful localization of the watering can through 3D instance segmentation.
    Following this, an optimal robot positioning is computed by the navigation planner and the object is grasped.
    The right side of the figure details the drawer detection and manipulation process.
    Multiple images are captured for robust detection.
    Subsequently, the robot is maneuvered into position to facilitate drawer opening. 
    This dual-phase approach demonstrates the integration of object detection, navigation planning, and execution within a dynamic scene.
    On the respective sides we illustrate example objects and handles in various levels of difficulty.
    }
    \label{fig:spot_highlighting_and_navigation}
    \vspace{-10pt}
\end{figure*}

\textbf{3D instance segmentation.}
Given a 3D point cloud as input, 3D segmentation assigns each point in a scene to an object mask and semantic label \cite{he2017mask, genova2021learning, schult2023mask3d, sun2023superpoint, Huang2023Segment3D, Kreuzberg22ECCVW, lei2020spherical, cciccek20163d, li2018pointcnn, wang2019associatively}.
Instance segmentation additionally distinguishing between different object instances. For example, Mask3D \cite{schult2023mask3d} relies on a Transformer-based architecture to directly predict 3D instance masks from point clouds, by learning instance queries through iterative attention to multi-scale point cloud features.
More recently, these works were extended to \emph{open-vocabulary} 3D segmentation which enables searching a 3D scene for arbitrary object instances as described by a natural language query \cite{takmaz2023openmask3d, Xu2023OpenVocabularyPS, engelmann2023opennerf, Chen2023OpenvocabularyPS, Peng2023OpenScene}. This is enabled by features obtained from visual-language models (VLM) such as CLIP\cite{radford2021learning}, SigLIP\cite{zhai2023sigmoid} or SILC\cite{naeem2023silc}. 
In this work, we rely on OpenMask3D \cite{takmaz2023openmask3d} to localize arbitrary objects for the robot to interact with (\emph{e.g.,} ``watering can", see Fig.\ref{fig:spot_highlighting_and_navigation}).    

\begin{figure}
    \centering
    \includegraphics[width=\linewidth]{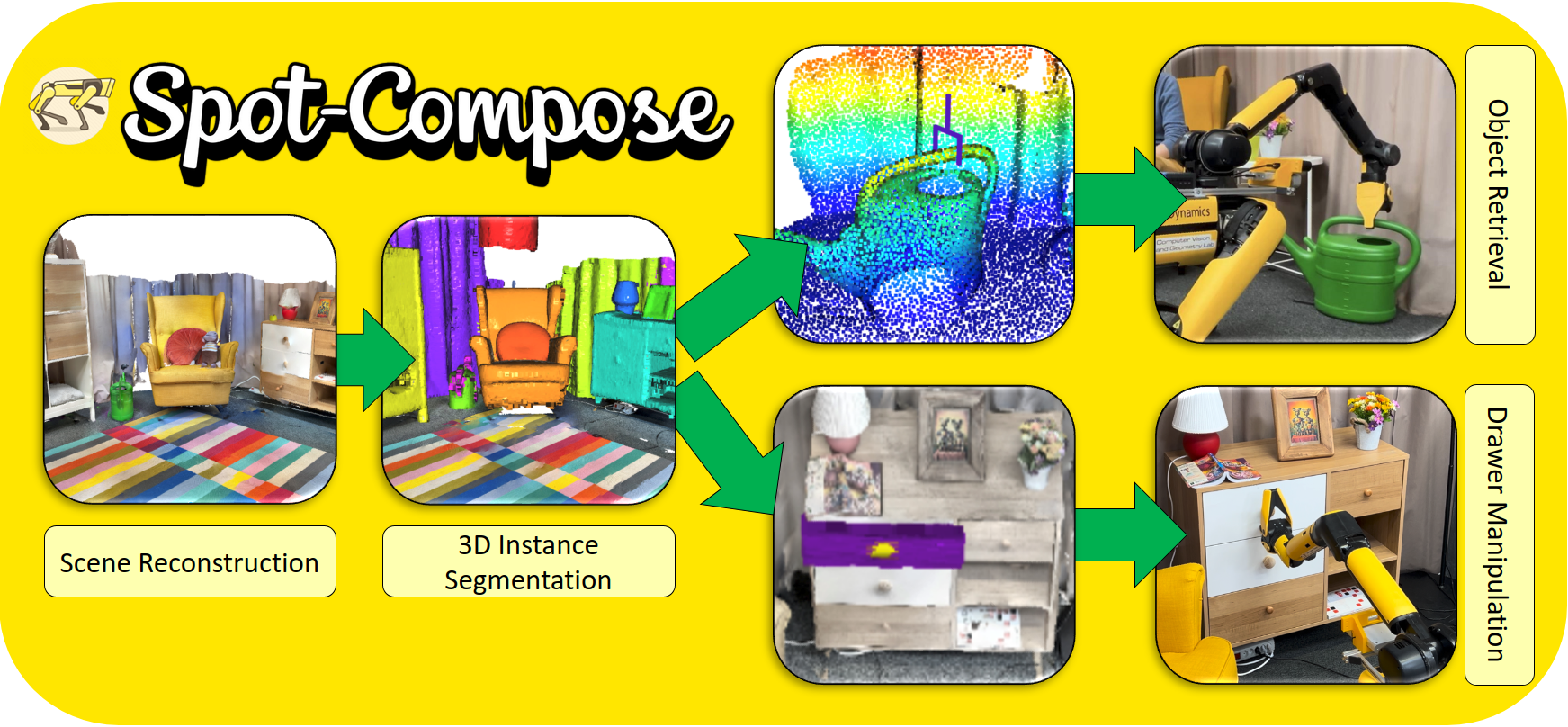}
    \caption{\textbf{Overview of the Spot-Compose pipeline.}
    Given a previously acquired point cloud, we segment the scene and localize the wanted object via a natural language query.
    For object retrieval (top), we isolate the object to determine the most effective grasp.
    For drawer manipulation (bottom) we use the cabinet position to point our camera for 2D drawer detection.
    }
    \vspace*{-12pt}
    \label{fig:paper_appetizer}
\end{figure}

\textbf{Grasp pose estimation in point clouds.}
Given an object, robotic grasping determines the most effective ways for robotic two-finger grippers to grasp objects in various situations \cite{kumra2020antipodal, zeng2022robotic, hundt2019costar, karunratanakul2020grasping, ainetter2021end, chu2018real,zurbrugg2024icgnet}.
Early works \cite{ten2017grasp,mousavian20196,duan2021robotics} propose sampling-based grasp estimation, which typically require longer inference times.
More recent approaches experiment with end-to-end learning to address this issue \cite{fang2020graspnet, hoang2022voting}.
In this work, we rely on AnyGrasp~\cite{fang2023anygrasp} to predict two-finger grasps directly on 3D point clouds.
It utilizes a dense supervision strategy combining real perception and analytic labels in the spatial-temporal domain, including awareness of the object's center-of-mass for improved grasp stability. It also models environment awareness, which simplifies grasp estimation and avoids collisions with the background and other objects.

\textbf{Robot task planning.}
Robot task planning refers to the process of creating and organizing a series of actions for a robot to achieve specific goals \cite{hundt2019costar, rosinol20203d, chang2023goat, ahn2022can}.
``\textit{ASC: Adaptive Skill Coordination for Robotic Mobile Manipulation}" \cite{yokoyama2023adaptive} presents a method for performing long-horizon tasks like mobile pick-and-place using three key components: a library of basic visual-motor skills, a skill coordination policy, and a corrective policy for adapting skills in novel situations.
\textit{OK-Robot} \cite{liu2024ok} presents an open knowledge-based robotics framework integrating VLMs with navigation and grasping primitives for efficient pick-and-drop operations in home environments. 
Unlike \textit{OK-Robot}, we employ modern 3D segmentation techniques and directly rely on the 3D scene representation, and demonstrate interactions with articulated objects such as opening drawers.

\section{TECHNICAL COMPONENTS AND METHOD}
Our objective is twofold: firstly, to enable the picking of objects across a wide range of shapes and sizes;
and secondly, to facilitate access to concealed spaces, defined as areas that become accessible only by manipulating elements such as drawers or doors, and are otherwise inaccessible or hidden from view.
Towards that goal, we require a variety of capabilities, which are
(A) semantic point cloud segmentation to identify objects that can be interacted with, 
(B) point cloud-dependent grasp pose estimation, 
(C) adaptive navigation, and 
(D) dynamic drawer detection and subsequent estimation of the axis of motion. 
Finally, in (E) we will provide a brief overview of additional functionalities that can be readily implemented with the aforementioned skills.

In all our experiments, we utilize the Boston Dynamics Spot \cite{bostondynamic2024spot} robot. 
Our framework builds upon the official SDK \cite{bostondynamic2024spotsdk} and is modular in nature to easily allow for employment of more recent methods as they become available.

\subsection{Object localization via 3D instance segmentation}
\label{sec:obj_local}
One key element of our approach is the open-vocabulary 3D instance segmentation model, which we employ to map and understand the 3D space and then localize any object of interest specified through a natural language query.

\textbf{Acquisition of 3D environmental data.}
We assume the availability of a pre-scanned 3D representation of the environment.
These representations can be obtained through various methods, including the use of modern smartphones equipped with LiDAR scanners and an associated scanning apps.
We conduct all of our environment scans using an iPhone 13 Pro Max using the \textit{3D Scanner App} \cite{3dscannerapp2023}.

\textbf{Open-vocabulary 3D Instance Segmentation.}
Contrary to previous approaches, recent developments in 3D instance segmentation enable the localization of any specified object directly from the 3D point cloud data.
This represents a significant advancement over traditional 2D instance segmentation by facilitating the extraction of a detailed, object-specific mask at the point level.
We propose, that utilizing this mask should aid in subsequent planning and picking processes, allowing for more precise object manipulation.

In this paper, we deploy OpenMask3D \cite{takmaz2023openmask3d}, which on top of instance segmentation, allows querying of segmentation masks based on natural language input.
This enhances the robot by allowing intuitive command input in natural language, broadening accessibility and user-friendliness.

\subsection{Adaptive grasping}
\label{sec:obj_dep_grasp}
The crucial stage in robotic object manipulation lies in the grasp pose estimation.
We implement the AnyGrasp system \cite{fang2023anygrasp} for this purpose. Fang \emph{et al.}\cite{fang2023anygrasp} describe AnyGrasp as a ``unified system for fast, accurate,
7-DoF and temporally-continuous grasp pose detection, using a parallel gripper".
Additionally, this method is aware of the object's center of gravity and its environment, filtering grasps that would be impossible to execute due to surrounding obstacles.

\textbf{Inference.}
By  default, AnyGrasp is tuned to identify grasp poses based on the frontal view of the given point cloud.
To expand our grasp detection capabilities and encompass all potential grasp poses, we run  multiple detection iterations, each with distinct initial rotations of the object.
For each perspective, we obtain the top $k$ grasps for post-processing.
Subsequently, we filter the poses based on a set of criteria, namely (1) have a positive confidence score and (2) be located on the object point cloud.

\subsection{Adaptive navigation and joint optimization}
\label{sec:ada_nav}
Before grasping, we need to decide where to position the robot, such that it has a good vantage point.
For this, we sample a set of positions radiating outward from the grasp item.
For each position, we check whether it (1) lies within the scene and (2) has a direct line of sight to the object.
Over the remaining body and grasp poses we decide on the best combination via joint optimization.

\begin{equation}
    s = s_{\text{grasp}} + \lambda_{\text{body}} \cdot s_{\text{body}} + \lambda_{\text{align}} \cdot s_{\text{align}}
\end{equation}

where, $s_{\text{grasp}}$ is the confidence score of the grasp, while 
\begin{equation}\label{eq:body_planning}
    s_{\text{body}} = d_{\text{obstacles}} - \lambda_{\text{item}} d_{\text{item}}
\end{equation}
defines a metric on the body pose, such that
$d_{\text{obstacles}}$ denotes the distance to the nearest non-grasp item object in the environment, while $d_{\text{item}}$ denotes the distance to the grasp item.
By adjusting $\lambda_{\text{item}}$, this metric strikes a good trade-off between avoiding collisions with the environment, while staying close enough to the object to allow for easy grasping.

Finally, $s_{\text{align}}$ represents
\begin{equation}
    s_{\text{align}} = \tanh\Bigl(
    T \cdot
    \frac{\Vec{x}_{\text{rt}}}{||\Vec{x}_{\text{rt}}||}
    \cdot
    \frac{\Vec{x}_{\text{g}}}{||\Vec{x}_{\text{g}}||}
    \Bigr),
\end{equation}
\emph{i.e.}, the dot product between the normalized vectors $\Vec{x}_{\text{rt}}$, pointing from robot to target, and $\Vec{x}_{\text{g}}$, pointing in the grasp direction, scaled by some temperature $T$.
This term encourages body and grasp positions to be aligned with each other.
We add a tanh term to allow for slight misalignments.

We experimentally set $\lambda_{\text{body}} = 0.01$, $\lambda_{\text{align}} = 0.02$, $\lambda_{\text{item}} = 0.5$, and $T=1$. 
Note that not all scores $s$ have the same initial magnitude.
This configuration enables the model to focus on the following aspects in order of importance:
(1) finding the best grasp, 
(2) choosing a body position best aligned with that grasp, and
(3) choosing a body position distanced from any obstacles.
We have found these parameters to work well for our environment, however encourage experimentation with these values for new locations.

\subsection{Dynamic drawer detection and motion estimation}
\label{sec:draw_det}
The second key capability we introduce involves the dynamic detection and manipulation of drawers, comprising three subtasks: drawer and handle detection, estimation of the axis of motion, and actual grasp planning.
This skill is significant, enabling the robot to access spaces that are typically concealed, such as when searching for objects inside drawers.

\textbf{Drawer and handle detection.}
To initially identify all cabinets within the environment, we apply the method outlined in Section \ref{sec:obj_local}.
However, the detection of individual drawers and associated handles is much less explored \cite{delitzas2024scenefun3d}.
The lack of distinct geometric features in drawers and insufficient resolution of the point cloud render it ineffective for accurately segmenting handles.
Instead, we opt to leverage the RGBD camera embedded in the robot's gripper to capture images of the cabinet and localize handles in the RGB image using 2D object detection techniques.
The final handle pose is computed by backprojecting into 3D using the associated depth value.
For 2D handle detection, we finetune a YOLOv8 model \cite{ultralytics2023yolo} on the DoorDetect dataset \cite{arduengo2021doordetect}.
To increase the robustness of our detection, we capture multiple images, utilizing Iterative Farthest Point Sampling (see Fig. \ref{fig:spot_highlighting_and_navigation}).

\textbf{Axis of motion estimation.}
A crucial aspect of the drawer pulling action is determining the axis of motion, which dictates the direction in which the drawer must be opened.
We find that this motion typically aligns with the normal of the drawer front.
To identify 3D points related to the drawer front, one might consider selecting points within a specific constant offset from the handle detection.
However, this approach fails to generalize effectively due to the variable distance to the camera and the drawer's shape.
Instead, we leverage the fact that our model successfully identifies both drawers and handles, and employ Hungarian Matching to pair the two detections. The matching cost is defined as
\begin{equation}
C_{\text{Hungarian}}(i,j) = -\bigl( \kappa \cdot \text{IoA}(h_i, d_j) + \text{Conf}(d_j)\bigr),
\end{equation}
where $h_i$ and $d_j$ denote the $i^{\text{th}}$ handle and the $j^{\text{th}}$ drawer instance, respectively. Here, Conf$(d_j)$ is the confidence score of the drawer prediction, and IoA$(h_i, d_j)$ (``Intersection over Area"), represents the proportion of the handle's bounding box that overlaps with the drawer's bounding box,
\begin{equation}
\text{IoA}(h_i, d_j) = \frac{\text{Intersection}(h_i, d_j)}{\text{Area}(h_i)}.
\end{equation}
This cost function is prioritizes handle bounding boxes that fall within the drawer bounding boxes, using the confidence of the drawer detection as a secondary criterion. The constant $\kappa = 10$ balances the significance of these two factors.

To estimate the axis of motion, we project 3D depth capture into the image, select all points that lie within the drawer's, but not the handle's, bounding box, and employ RANSAC \cite{fischler1981ransac} plane estimation.

\begin{figure*}[!t]
\vspace*{-16pt}
\hspace*{-12pt}
    \centering
    \begin{tabular}{cccc}
        \rotatebox[origin=c]{90}{Grasping Experiment} \hspace*{-12pt} & 
        \makecell{\includegraphics[width=0.48\linewidth]{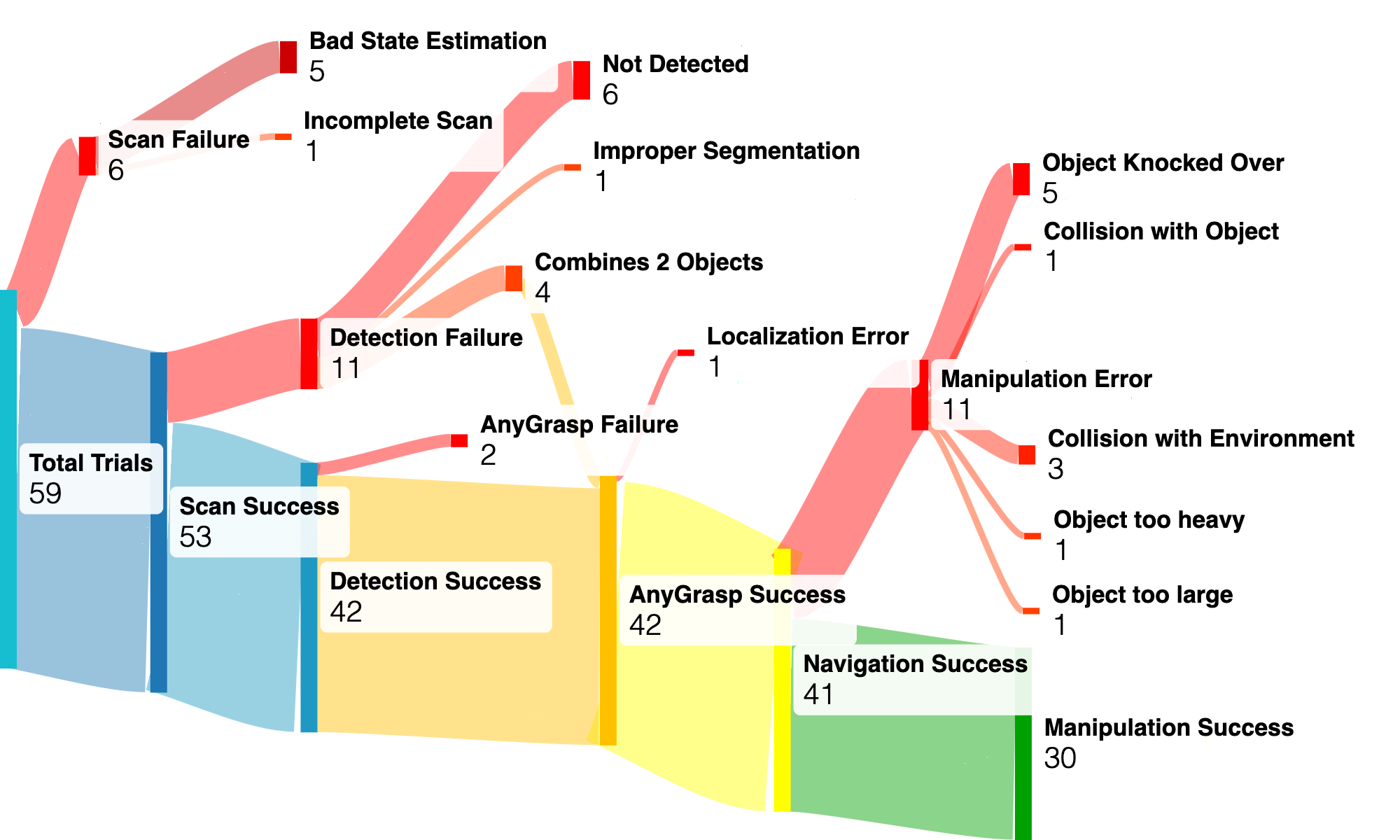}} & \hspace*{-12pt}
        \rotatebox[origin=c]{90}{Search Experiment} 
        \hspace*{-12pt} & 
        \makecell{\includegraphics[width=0.48\linewidth]{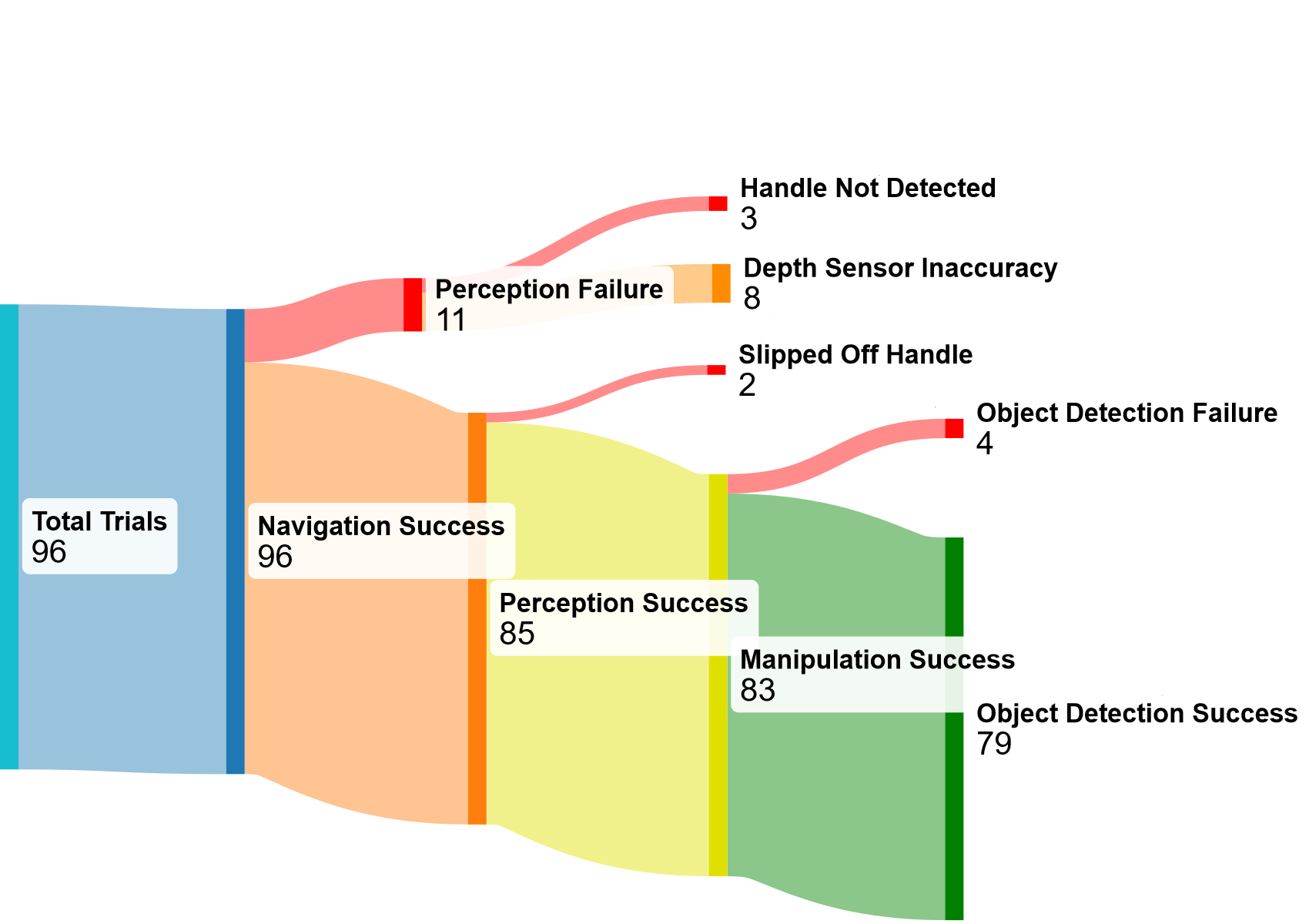}} \\
    \end{tabular}
    \vspace*{-8pt}
    \caption{\textbf{Grasping experiment results.}
    To evaluate the grasping capability of our framework, we conduct 59 trial runs across six different scenes and with 13 distinct objects.
    The test includes items and placements of varying difficulty.
    We observed an overall success rate of 51\%, with the highest failure rate occurring in detection and manipulation.
    \textbf{Search experiment results.}
    For this evaluation, we conduct 16 runs with six individual drawers.
    Moreover, we explore combinations of 15 handles and 19 objects, experimenting with various pairings.
    We observe a 82\% success rate, with the majority of failure cases being connected to bad perception, especially an inaccurate Time-of-Flight depth sensor.
    }
    \vspace{-10pt}
    \label{fig:experiments}
\end{figure*} 
\textbf{Refinement and impedance-based pulling.}
After estimating all the necessary components for the pulling motion, the remaining step is execution.
We position the robot at a predetermined distance in front of the cabinet, aligning it parallel to the axis of motion.
Given that our predictions are unlikely to be perfectly accurate, we implement two additional refinements.
The first involves capturing another image in front of the drawer to refine both the coordinate and the axis of motion from a closer proximity.
Secondly, an impedance-based pulling motion allows the robot some flexibility in directions orthogonal to the axis of motion.

\subsection{Expansion of capabilities}
\label{sec:other_caps}
With the foundational framework established, extending the system to incorporate additional functionalities becomes a relatively straightforward task.

\textbf{Playing fetch.}
Combining an appropriate human detection model, such as Human3D~\cite{takmaz20233d} or VitPose~\cite{xu2022vitpose}, in combination with the object retrieval function described in Section \ref{sec:obj_dep_grasp}, the robot can pick up and deliver objects to humans.

\textbf{Search.}
The ability to open and close drawers is only one building block.
When this functionality is integrated with the camera located on the end-effector, alongside a contemporary open-vocabulary object detection approach (such as OWLv2 \cite{minderer2023scaling}), it becomes straightforward to develop a mobile search robot capable of exploring concealed areas.

\section{EXPERIMENTS}
We assess our framework on two separate experiments: (1) grasping, and (2) search.
The experiments are designed to evaluate the high level functions of our model, while highlighting multiple the facets of the framework.

\textbf{Grasping.}
To test the dynamic grasping ability of our model, we evaluate it on a set of 6 different scenes with a varying difficulty levels of both object graspability and placement.
For an example of different object difficulties, please refer to Fig. \ref{fig:spot_highlighting_and_navigation}.
The level of difficulty was decided based on deformability of the object and amount of possible grasps.
All objects must be placed in a free-standing manner.
Each run is repeated once, to test for robustness and our results are illustrated in Fig. \ref{fig:experiments} and Table \ref{tab:conf_grasp}.
We observe an overall success rate of 51\%, where the ease of grasping a respective object is the main predictor of a successful overall grasp.
While we are able to manipulate even very difficult objects, such as watering cans, and navigate difficult locations, robustness in these cases tends to suffer.

\begin{table}[h]
\vspace{-6pt}
    \centering
    \caption{\textbf{Success Rate Breakdown: Object vs. Placement Difficulty.}
    We illustrate the manipulation success rate across different difficulty levels of objects and placements, observing an inverse correlation with higher difficulty.
    }
    \vspace{-3pt}
    \begin{tabular}{|l|c|c|c|}
        \hline
        \diagbox{\textbf{Object}}{\textbf{Placement}} & \textbf{Easy} & \textbf{Medium} & \textbf{Hard} \\ \hline
        \textbf{Easy}   & 75\% & 100\% & 100\% \\ \hline
        \textbf{Medium} & 90\% & 75\% & 50\% \\ \hline
        \textbf{Hard}   & 83\% & 25\% & 40\% \\ \hline
    \end{tabular}
    \label{tab:conf_grasp}
    \vspace*{-6pt}
\end{table}

\textbf{Search.}
We aim to evaluate the drawer localization and 2D detection capabilities of our approach.
The setup is as follows:
We initialize the robot with the position of the two cabinets derived via instance segmentation of the environment, as well as a natural language search term.
Subsequently, the robot is tasked with detecting all drawers within the cabinets, opening them to inspect their contents, and identifying any items within.
Should the sought-after object be found among the contents, the robot must record the drawer in which the object was located in 3D space.

The results are illustrated in Fig \ref{fig:experiments}.
We report an 82\% success rate, with the majority of failure cases being related to inaccurate depth sensing.
By capturing multiple perspectives of a given cabinet, we are able to robustly detect even small, color-matching, or oddly-shaped handles in the vast majority of cases.
However, this is at the cost of additional inference time, representing a trade-off for real-work applications.

\textbf{Inference times.}
We list the expected inference times for localization (0.221s), one-time 3D instance segmentation (271s) grasp pose estimation (13.7s), navigation planning (24.0s), joint optimization (0.3ms), drawer detection (0.84s) and zero-shot object detection (2.85s).
All times except for localization and optimization include latency, as they are executed on an external NVIDIA RTX 4090 GPU.

\section{CONCLUSION}
In this paper, we have presented a comprehensive framework to efficiently build new functionality for the Spot robot, increasing accessibility for researchers beyond the field of robotics.
We utilize it to enhance the capabilities of robots to interact within human-centric environments, specifically through dynamic object retrieval and drawer manipulation tasks.
Our work leverages state-of-the-art techniques in 3D instance segmentation, grasp pose estimation, and object detection to enable mobile manipulation in real-world settings.

The experiments underline the practical applicability of our approach, showcasing the potential for robots to perform complex tasks in environments designed for humans.
Finally, the modular nature of our framework allows for the seamless integration of future advancements in perception and manipulation technologies.

Current limitations of our approach include long latencies arising from 3D instance segmentation and 360$^\circ$ grasp pose estimation.
In future work, we plan to focus on efficient grasp trajectory planning and more sophisticated joint optimization to enhance the robustness of our model.
We encourage everybody to build on top of our framework, enabling more streamlined implementation and sophisticated actions.

\clearpage
\bibliography{references}

\end{document}